\documentclass{article} 
\pdfoutput=1
\usepackage{nips13submit_e,times}
\usepackage{hyperref}
\usepackage{url}
\usepackage{graphicx}
\usepackage{booktabs}

\title{Face Detection using Deep Learning:\\An Improved Faster RCNN Approach}

\author{
Xudong Sun, Pengcheng Wu, Steven C.H. Hoi\\
DeepIR Inc.\\
\texttt{\{sunxd,tim,steven\}@deepir.com} \\
}

\nipsfinalcopy 

\begin{document}

\maketitle

\begin{abstract}

In this report, we present a new face detection scheme using deep learning and achieve the state-of-the-art detection performance on the well-known FDDB face detetion benchmark evaluation. In particular, we improve the state-of-the-art faster RCNN framework by combining a number of strategies, including feature concatenation, hard negative mining, multi-scale training, model pretraining, and proper calibration of key parameters. As a consequence, the proposed scheme obtained the state-of-the-art face detection performance, making it the best model in terms of ROC curves among all the published methods on the FDDB benchmark.
\end{abstract}

\section{Introduction}

Face detection is a fundamental and important problem in computer vision and pattern recognition, which has been widely studied over the past few decades. Face detection is one of the important key steps towards many subsequent face-related applications, such as face verification\cite{sun2014deep,taigman2014deepface}, face recognition \cite{kumar2011real,parkhi2015deep,schroff2015facenet}, and face clustering \cite{schroff2015facenet}, etc. Following the pioneering work of Viola–Jones object detection framework \cite{viola2001rapid,viola2004robust}, numerous methods have been proposed for face detection in the past decade. Early research studies in the literature were mainly focused on extracting different types of hand-crafted features with domain experts in computer vision, and training effective classifiers for detection and recognition with traditional machine learning algorithms. Such approaches are limited in that they often require computer vision experts in crafting effective features and each individual component is optimized separately, making the whole detection pipeline often sub-optimal.

In recent years, deep learning methods, especially the deep convolutional neural networks (CNN), has achieved remarkable successes in various computer vision tasks, ranging from image classification to object detection and semantic segmentation, etc. In contrast to traditional computer vision approaches, deep learning methods avoid the hand-crafted design pipeline and have dominated many well-known benchmark evaluations, such as ImageNet Large Scale Visual Recognition Challenge (ILSVRC) \cite{krizhevsky2012imagenet}. Along with the popularity of deep learning in computer vision, a surge of research attention has been emerging to explore deep learning for resolving face detection tasks.

In general, face detection can be considered as a special type of object detection task in computer vision. Researchers thus have attempted to tackle face detection by exploring some successful deep learning techniques for generic object detection tasks. One of very important and highly successful framework for generic object detection is the region-based CNN (RCNN) method \cite{girshick2014rich}, which is a kind of CNN extension for solving the object detection tasks. A variety of recent advances for face detection often follow this line of research by extending the RCNN and its improved variants.




Following the emerging trend of exploring deep learning for face detection, in this paper, we propose a new face detection method by extending the state-of-the-art Faster R-CNN algorithm \cite{ren2015faster}. In particular, our scheme improves the existing faster RCNN scheme by combining several important strategies, including feature concatenation \cite{bell2015inside}, hard negative mining, and multi-scale training, etc. We conducted an extensive set of experiments to evaluate the proposed scheme on the well-known Face Detection Dataset and Benchmark (FDDB)\cite{jain2010fddb}, and achieved the state-of-the-art performance (ranking the best among all the published approaches).

The rest of this report is organized as follows. Section 2 briefly reviews the related work in face detection literature and recent advances of deep learning approaches. Section 3 presents the proposed deep learning approach for face detection. Section 4 discusses our experiments and empirical results. Section 5 concludes this work.


\section{Related Work}

Face detection has extensively studied in the literature of computer vision. Before 2000, despite many extensive studies, the practical performance of face detection was far from satisfactory until the milestone work proposed by Viola and Jones \cite{viola2001rapid}\cite{viola2004robust}. In particular, the VJ framework \cite{viola2001rapid} was the first one to apply rectangular Haar-like features in a cascaded Adaboost classifier for achieving real-time face detection. However, it has several critical drawbacks. First of all, its feature size was relatively large. Typically, in a $24\times24$ detection window, the number of Haar-like features was 160,000\cite{viola2004robust}. In addition, it is not able to effectively handle non-frontal faces and faces in the wild.

To address the first problem, much effort has been devoted to coming up with more complicated features like HOG\cite{zhu2006fast}, SIFT, SURF\cite{li2013learning} and ACF\cite{yang2014aggregate}. For example, in \cite{liao2016fast}, a new type of feature called NPD was proposed, which was computed as the ratio of the difference between any two pixel intensity values to the sum of their values. Others aimed to speed up the feature selection in a heuristic way\cite{pham2007fast}\cite{brubaker2008design}. The well known Dlib C++ Library\cite{dlib09} took SVM as the classifier in its face detector. Other approaches, such as random forest, have also been attempted.

Enhancing the robustness of detection was another extensively studied topic. One simple strategy was to combine multiple detectors that had been trained separately for different views or poses\cite{jones2003fast}\cite{li2002statistical}\cite{huang2007high}. Zhu et al.\cite{zhu2012face} applied multiple deformable part models to capture faces with different views and expressions. Shen et al.\cite{shen2013detecting} proposed a retrieval-based method combined with discriminative learning. Nevertheless, training and testing of such models were usually more time-consuming, and the boost in detection performance was relatively limited. Recently, Chen et al.\cite{chen2014joint} constructed a model to perform face detection in parallel with face alignment, and achieved high performance in terms of both accuracy and speed.

Recent years have witnessed the advances of face detection using deep learning methods, which often outperform traditional computer vision methods significantly. For example, Li et al.\cite{li2016face} presented a method for detecting faces in the wild, which integrates a ConvNet and a 3D mean face model in an end-to-end multi-task discriminative learning framework. Recently, \cite{jiang2016face} applied the Faster R-CNN \cite{ren2015faster}, one of state-of-the-art generic object detector, and achieved promising results. In addition, much work has been done to improve the Faster R-CNN architecture. In \cite{qin2016joint}, joint training conducted on CNN cascade, region proposal network (RPN) and Faster R-CNN has realized end-to-end optimization. Wan et al.\cite{wan2016bootstrapping} combined Faster R-CNN face detection algorithm with hard negative mining and ResNet and achieved significant boosts in detection performance on face detection benchmarks like FDDB. In this work, we propose a new scheme for face detection by improving the Faster RCNN framework.

\section{Our Approach}

\subsection{Overview of Methodology}

\begin{figure}[!htbp]
  \centering
  \includegraphics[width=0.5\textwidth]{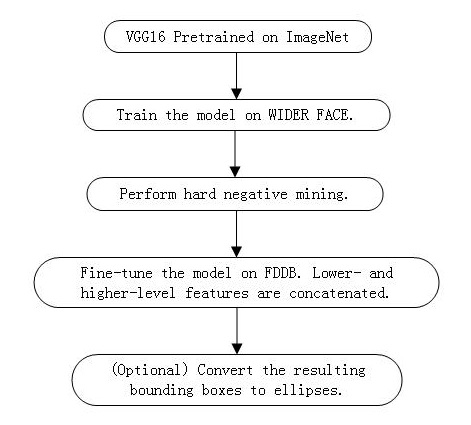}
  \caption{Flowchart of the training procedure of the proposed deep learning scheme}
  \label{fig:flowchart}
\end{figure}

Our method follows the similar deep learning framework of Faster RCNN, which has been shown to be a state-of-the-art deep learning scheme for generic object detection \cite{ren2015faster}. It essentially consists of two parts: (1) a Region Proposal Network (RPN) for generating a list of region proposals which likely contain objects, or called regions of interest (RoIs); and (2) a Fast RCNN network for classifying a region of image into objects (and background) and refining the boundaries of those regions. In this work, we propose to extend the Faster RCNN architecture for face detection, and train our face detection model by following the proposed procedure as shown in Figure \ref{fig:flowchart}.


First of all, we train the CNN model of Faster RCNN using the WIDER FACE dataset \cite{yang2016wider}. We further use the same dataset to test the pre-trained model so as to generate hard negatives. These hard negatives are fed into the network as the second step of our training procedure. The resulting model will be further fine-tuned on the FDDB dataset. During the final fine-tuning process, we apply the multi-scale training process, and adopt a feature concatenation strategy to further boost the performance of our model. For the whole training processes, we follow the similar end-to-end training strategy as Faster RCNN. As a final optional step, we convert the resulting detection bounding boxes into ellipses as the regions of human faces are more elliptical than rectangular.

In the following, we discuss three of the key steps in the proposed solution in detail.

\subsection{Feature Concatenation}

\begin{figure}[!htbp]
  \centering
  \includegraphics[width=\textwidth]{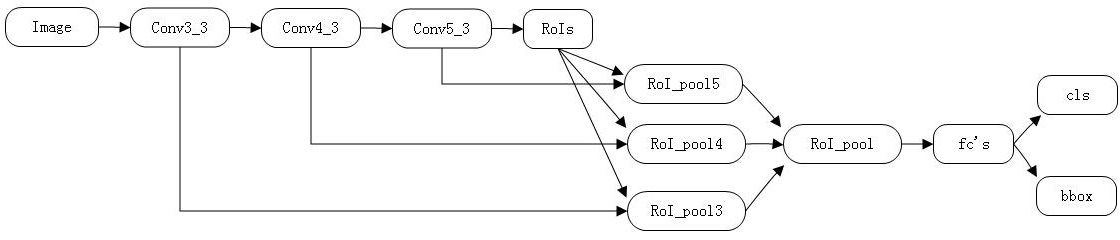}
  \caption{Network architecture of the proposed feature concatenation scheme}
  \label{fig:concat}
\end{figure}

For traditional Fast RCNN networks, the RoI pooling is performed on the final feature map layer to generate features of the region. Such an approach is not always optimal and sometimes may omit some important features, as features in deeper convolution layer output have wider reception fields, resulting in a grosser granularity. In the proposed solution, in order to capture more fine-grained details of the RoIs, we propose to improve the RoI pooling by combining the feature maps of multiple convolution layers, including both lower-level and high-level features. Inspired by the study in \cite{bell2015inside}, we propose to concatenate the pooling result of multiple convolutional feature maps to generate the final pooling features for detection tasks. Specifically, features from multiple lower-level convolution layers are ROI-pooled and L2-normalized, respectively. Those resulting features are then concatenated and rescaled to match the original scale of the features as if feature-concatenation had not been adopted. A 1x1 convolution is then applied to match the number of channels of the original network. The detailed architecture of this approach is shown in Figure \ref{fig:concat}.


\subsection{Hard Negative Mining}

Hard negative mining has been shown as an effective strategy for boosting the performance of deep learning, especially for object detection tasks including face detection \cite{wan2016bootstrapping}. The idea behind this method is that, hard negatives are the regions where the network has failed to make correct prediction. Thus, the hard negatives are fed into the network again as a reinforcement for improving our trained model. The resulting training process will then be able to improve our model towards fewer false positives and better classification performance.

In our approach, hard negatives were harvested from the pre-trained model from the first step of our training process. We then consider a region as hard negative if its intersection over union (IoU) over the ground truth region was less than 0.5. During the hard negative training process, we explicitly add those hard negatives into the RoIs for finetuning the model, and balance the ratio of foreground and background to be about 1:3, which is the same as the ratio that we use in the first step.

\subsection{Multi-Scale Training}

The Faster RCNN architecture typically adopt a fixed scale for all the training images. By resizing the images to a random scale, the detector will be able to learn features across a wide range of sizes, thus improving its performance towards scale invariant. In this work, we randomly assign one of three scales for each image before it is fed into the network. The details are given in our experimental section. Our empirical results show that the use of multi-scale training makes our model more robust towards different sizes, and improve the detection performance on benchmark results.

\section{Experiments}

\subsection{Experimental Setup}

We conduct an empirical study of evaluating the proposed face detection solution on the well-known FDDB benchmark testbed \cite{jain2010fddb}, which has a total of 5,171 faces in 2,845 images, including various detection challenges, such as occlusions, difficult poses, and low resolution and out-of-focus faces.

For implementation, we adopt the Caffe framework \cite{jia2014caffe} to train our deep learning models. VGG16 was selected to be our backbone CNN network, which had been pre-trained on ImageNet. For the first step, WIDER FACE training and validation datasets were selected as our training dataset. We gave each ground-truth annotation a difficulty value, according to the standard listed in Table \ref{tab:difficulty}. Specifically, all faces were initialized with zero difficulty. If a face was satisfied with a certain condition listed in Table \ref{tab:difficulty}, we add the corresponding difficulty value. We ignored those annotations whose difficulty values greater than 2. Further, all the images with more than 1000 annotations were also discarded.

\begin{table}[!htbp]
  \centering
  \caption{Difficulty Value Assignment Strategy}
  \label{tab:difficulty}
  \resizebox{\textwidth}{!}{
    \begin{tabular}{ccccccc}
      \toprule
      \multicolumn{2}{c}{Blur} & Expression & Illumination & \multicolumn{2}{c}{Occlusion} & Pose \\
      Normal Blur & Heavy Blur & Extreme Expression & Extreme Illumination & Partial Occlusion & Heavy Occlusion & Atypical Pose \\
      \midrule
      0.5 & 1 & 1 & 1 & 0.5 & 1 & 1 \\
      \bottomrule
    \end{tabular}
  }
\end{table}

The pre-trained VGG16 model was trained on this aforementioned dataset for 110,000 iterations with the learning rate set to 0.0001. During this training process, images were first re-scaled while always keeping the original aspect ratio. The shorter side was re-scaled to be 600, and the longer side was capped at 1000. Horizontal flipping was adopted as a data augmentation strategy. During the training, 12 anchors were used for the RPN part, which covers a total size of $64\times64$, $128\times128$, $256\times256$, $512\times512$, and three aspect ratios including 1:1, 1:2, and 2:1. After the non-maximum suppression (NMS), 2000 region proposals were kept. For the Fast RCNN classification part, an RoI is treated as foreground if its IoU with any ground truth is greater than 0.5, and background otherwise. To balance the numbers of foregrounds and backgrounds, those RoIs were sampled to maintain a ratio of 1:3 between foreground and background.

For the second step, the aforementioned dataset was fed into the network. Those output regions, whose confidence scores are above 0.8 while having IoU values with any ground-truth annotation are less than 0.5, were regarded as the ``hard negatives". The hard negative mining procedure was then taken for 100,000 iterations using a fixed learning rate of 0.0001, where those hard negatives were ensured to be selected along with other sampled RoIs. Finally, the resulting model was further fine-tuned on the FDDB dataset to yield our final detection model.

To examine the detection performance of our face detection model on the FDDB benchmark, we conducted a set of 10-fold cross-validation experiments by following the similar settings in \cite{jain2010fddb}. For each image, in addition to performing the horizontal flipping, we also randomly resize it before feeding it into the network. Specifically, we resize each image such that its shorter side will be one of $480, 600, 750$. Similar to the policy taken in the first step, we ensure that the longer side would not exceed 1250.

During the training process, we apply the feature concatenation strategy as introduced in the previous section. Specifically, we concatenated the features pooled from conv3\_3, conv4\_3, and conv5\_3 layers. As illustrated in \cite{bell2015inside}, the scale used after the features being concatenated could be either refined or fixed. Here we used a fixed scale of 4700 for the entire blob, both in the training and test phases. We fine-tuned the model for 40,000 iterations using a fixed learning rate of 0.001 to obtain our final models.

During the test phase, a query image was first re-scaled by following the same principle as in the first stage. For each image, a total of 100 region proposals were generated by the RPN network during the region proposal generation step. A selected region proposal would be regarded as a face if the classification confidence score is greater than 0.8. In our approach, the NMS threshold was set to 0.3. For the analysis purposes, we also output all the region proposals whose confidence scores are greater than 0.001 in our experiments.

\begin{figure}[!htbp]
  \centering
  \includegraphics[width=0.9\textwidth]{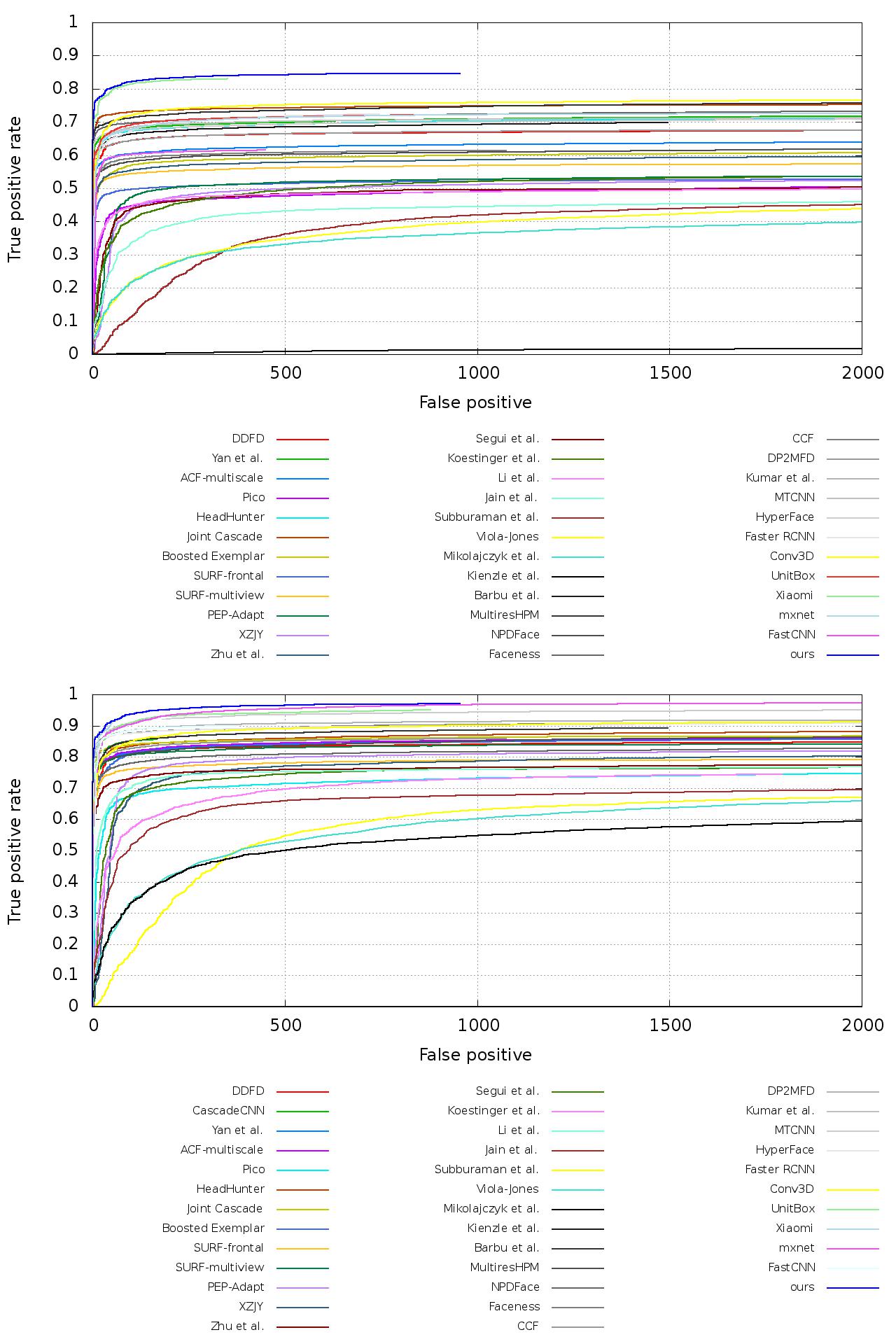}
  \caption{ROC curves of FDDB cross-validation result (top: continuous ROC result; bottom: discrete ROC result)}
  \label{fig:compareROC}
\end{figure}
\begin{figure}[htpb]
  \centering
  \vspace{-0.1in}
  \includegraphics[width=0.8\textwidth]{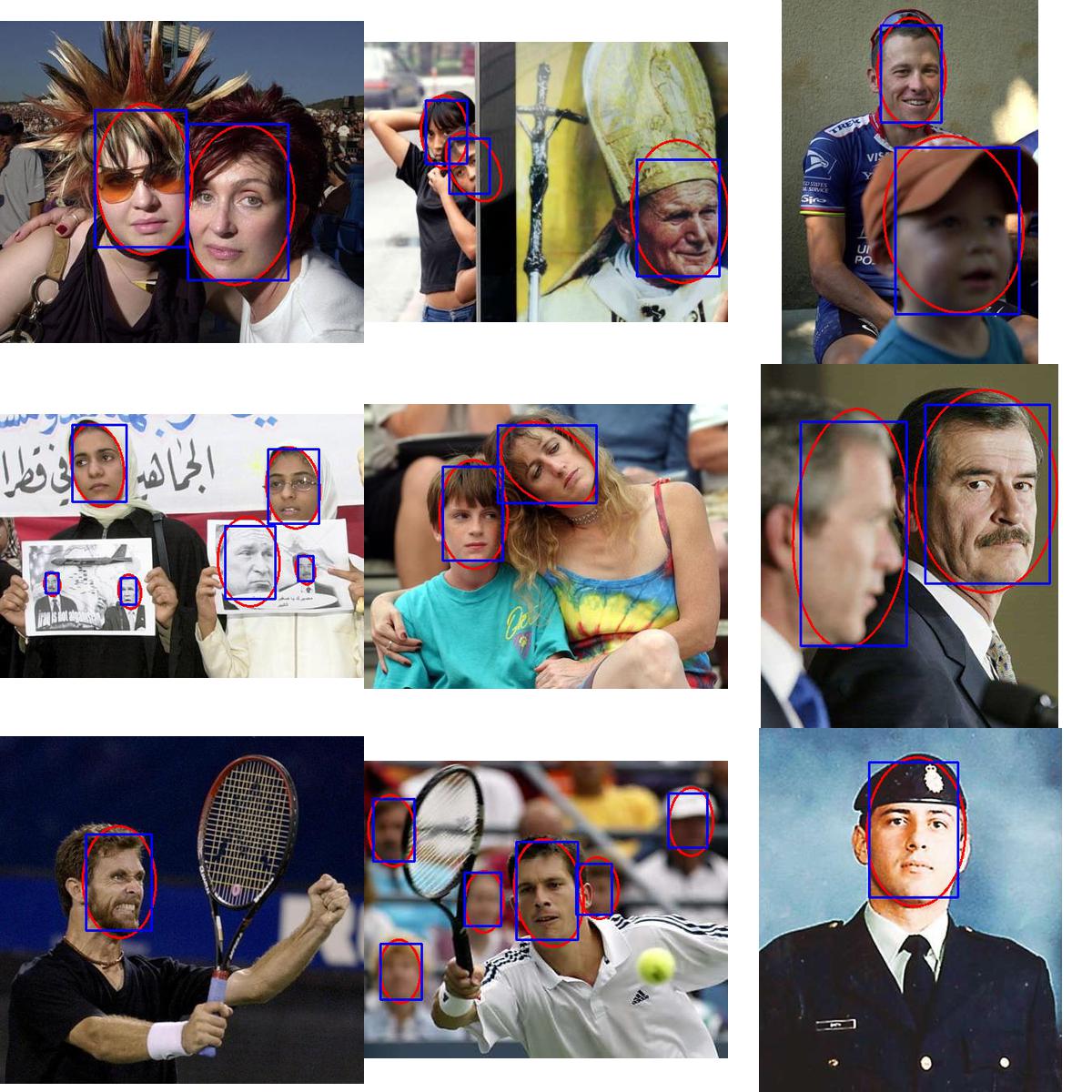}\vspace{-0.1in}
  \caption{Selected detected faces (red: annotation; blue: detection result)}
  \label{fig:detection}
  \vspace{-0.1in}
\end{figure}
\begin{figure}[htpb]
  \centering
  \includegraphics[width=0.9\textwidth]{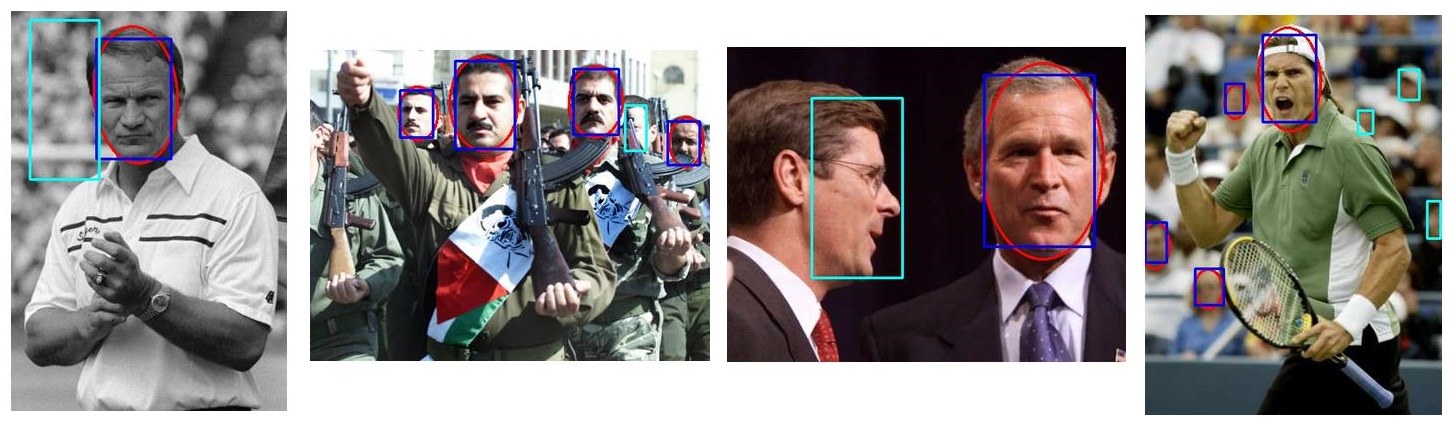}\vspace{-0.1in}
  \caption{Selected false positives of the proposed method on FDDB (red: annotation; blue: detection results; cyan: false positives)}
  \label{fig:fp}
  \vspace{-0.1in}
\end{figure}
\begin{figure}[htpb]
  \centering
  \includegraphics[width=0.9\textwidth]{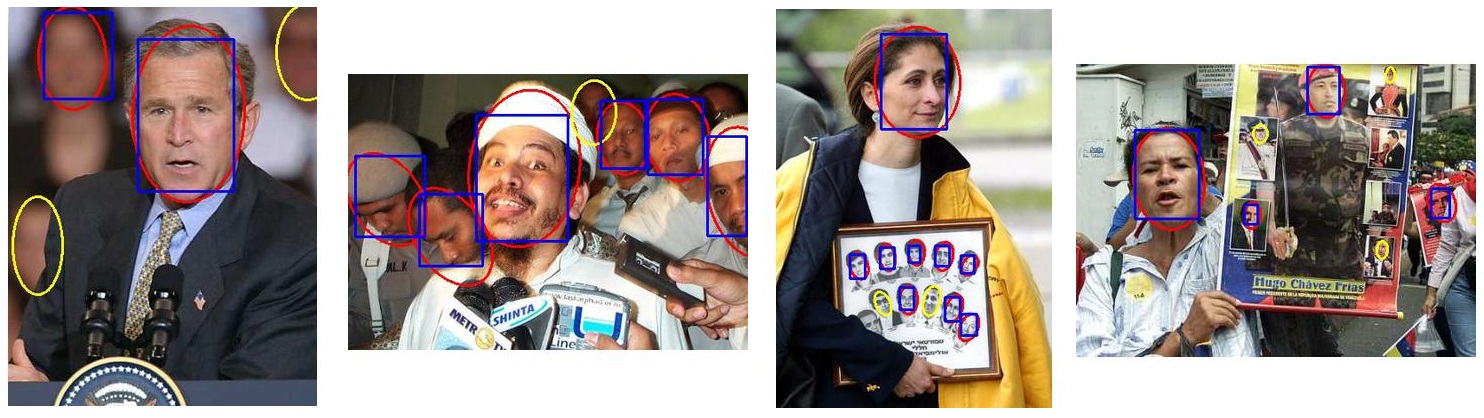}\vspace{-0.1in}
  \caption{Selected false negatives of the proposed method on FDDB  (red: annotation; blue: detection results; yellow: false negatives)}
  \label{fig:fn}
  \vspace{-0.1in}
\end{figure}

\subsection{Experimental Results on FDDB Benchmark}


Figure \ref{fig:compareROC} gives the detailed comparisons of two kinds of ROC scores for all the published methods submitted to the FDDB benchmark. Compared with the other results of the published methods, in terms of the standard ROC curves, the result obtained by our submitted model scores the highest among all the published methods, especially for the continuous ROC score where our method clearly  outperforms the second highest method \cite{wan2016bootstrapping}. The promising results validate the effectiveness of the proposed method for face detection using deep learning techniques.

\begin{figure}[!htbp]
  \centering
  \includegraphics[width=0.9\textwidth]{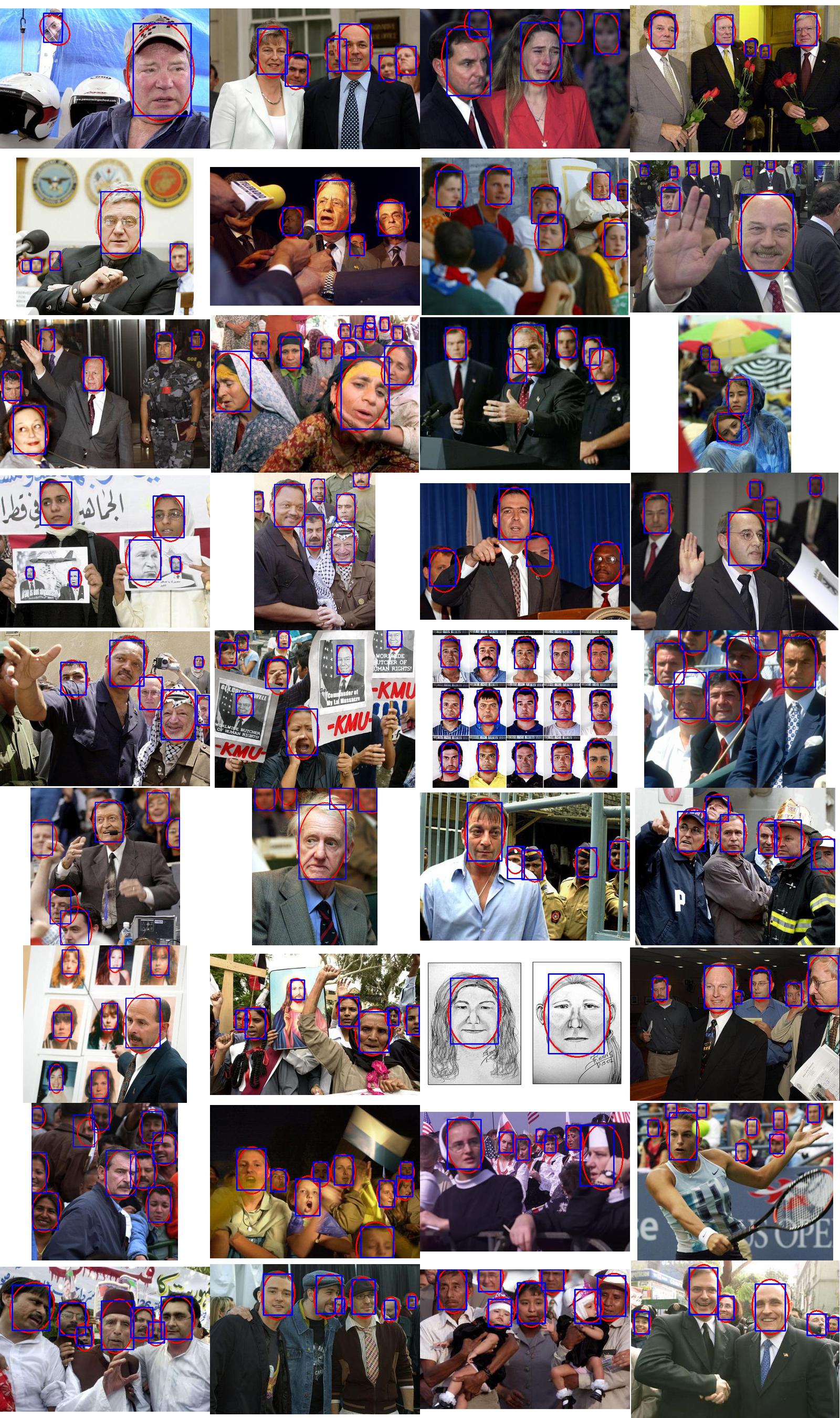}
  \caption{Curated examples on FDDB (red: annotation; blue: detection results)}
  \label{fig:curated}
\end{figure}

In addition to the quantitative evaluation results, we also randomly choose some qualitative results of face detection examples for different cases, as shown in Figure \ref{fig:detection}, Figure \ref{fig:fp}, and Figure \ref{fig:fn} (and more other examples in Figure \ref{fig:curated}). For example, Figure \ref{fig:detection} demonstrates that our model is able to detect some difficult cases, such as non-frontal faces, heavily occluded faces, faces with low resolution, and faces with extreme poses and/or illumination. Figure \ref{fig:fp} lists some selected false positives, where it seems that most of the false positives are actually missing annotations. Figure \ref{fig:fn} lists some of the false negatives, which includes some very challenging cases, such as blur faces, heavily occluded faces, and extremely small faces.

\subsection{Ablation Experiments}

\begin{figure}[htb]
  \centering
  \includegraphics[width=0.9\textwidth]{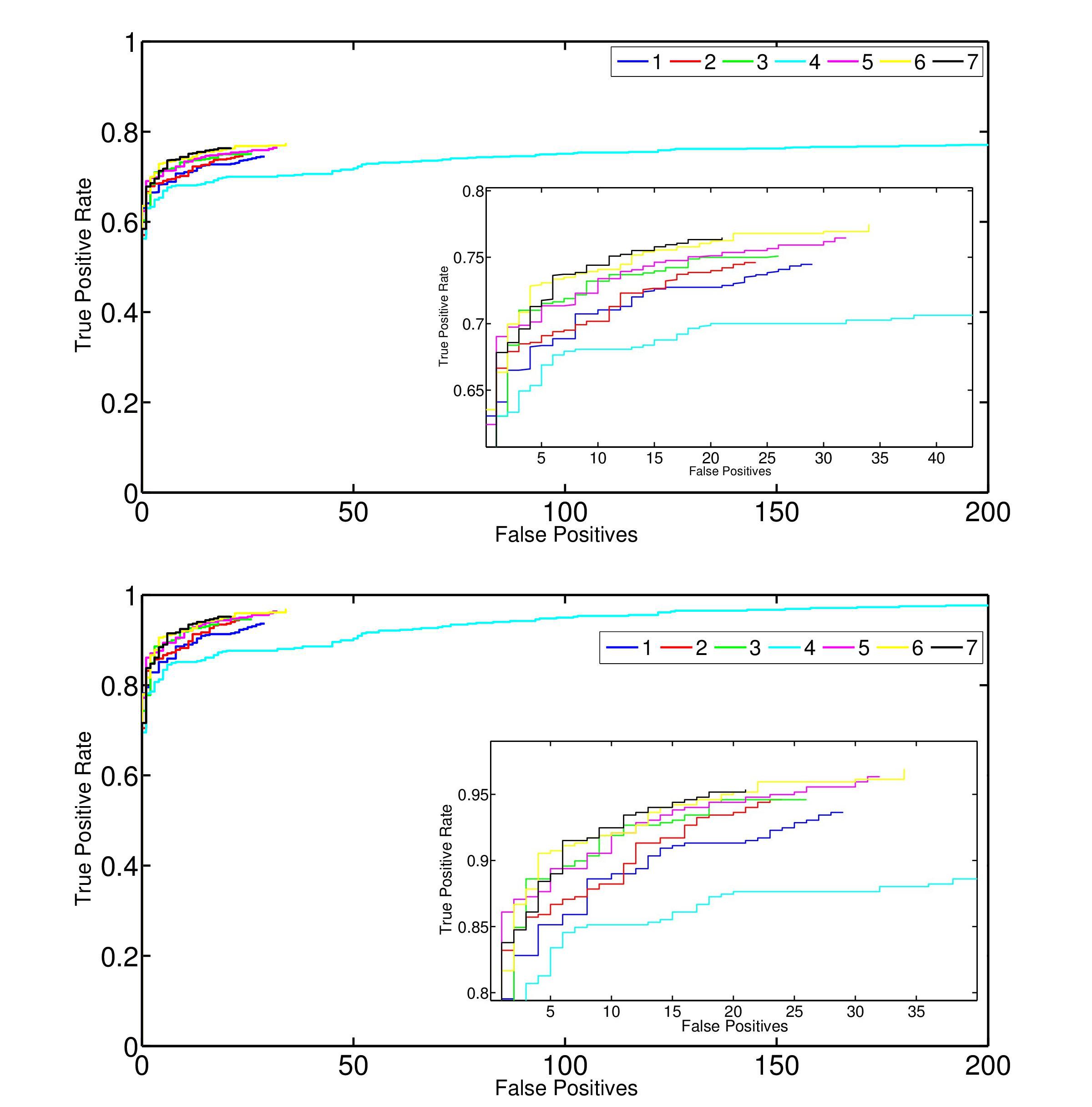}\vspace{-0.1in}
  \caption{Comparisons of Continuous ROC curves (top) and discrete ROC curves (bottom) for different experimental settings for our ablation studies. These experimental results shown here are only for the fold 7 of our cross-validation experiments; the other experimental folds are similar. The figures on the bottom right are magnified views of the selected regions. The detection bounding boxes are not converted to ellipses. (Best viewed in color).}
  \label{fig:methods}
\end{figure}

To further gain the deep insights of the improvements obtained by the proposed method, we conduct more additional experiments for ablation studies as listed in Table \ref{tab:experiments}, where we aim to examine the effectiveness and contributions of different strategies used in the proposed method. Figure \ref{fig:methods} shows the detailed experimental results of the ablation studies for examining several different settings.

\begin{table}[htb]
  \centering
  \caption{Additional experiments for ablation studies of the proposed solution}
  \label{tab:experiments}
  \resizebox{\textwidth}{!}{
    \begin{tabular}{cccccc}
      \toprule
      ID & \# Anchors & Train with WIDER FACE & Hard Negative Mining & Feature Concatenation & Multi-Scale Training \\
      \midrule
      1 & 9 & No & No & No & No \\
      2 & 12 & No & No & No & No \\
      3 & 12 & No & No & Yes & No \\
      4 & 12 & Yes & No & No & No \\
      5 & 12 & Yes & Yes & No & No \\
      6 & 12 & Yes & Yes & Yes & No \\
      7 & 12 & Yes & Yes & Yes & Yes \\
      \bottomrule
    \end{tabular}
  }
\end{table}

First of all, by examining the impact of anchor size, instead of using the default setting (9 anchors for RPN) by traditional faster RCNN, we compare this with our modification by adding a size group of $64\times64$, thus increasing the number of anchors to 12. Using this modification would allow our model to detect more small detection boxes (as shown in Experiment ID 1 vs ID 2).

Second, we examine the impact of pre-training on our model on additional larger-scale face data sets (such as WIDER FACE in our approach), since FDDB is a relatively small dataset (5171 faces in 2845 images).
However, the pre-training is not trivial as the WIDER FACE dataset is more challenging than FDDB, as it contains many difficult cases. As seen from experiment ID 4, although the detection recall was improved compared with Experiment ID 2, a simple training on WIDER FACE will yield many more false positives. By using the hard negative mining (as shown in Experiment ID 5), the number of false positives was reduced significantly.


Third, we examine the impact of feature concatenation strategy. As shown in our ablation study experiments (ID 2 vs ID 3, and ID 5 vs ID 6), feature concatenation turned out to be an effective strategy. By combining features from multiple layers, the model was able to learn features of multiple sizes, and was therefore better at classification.

Fourth, by further examining the impact of multi-scale training, we also observe a positive improvement from our ablation experiments (ID 6 vs ID 7). Specifically, by adopting the random scaling for data augmentation, the detection performance was further increased.

Finally, combining all the above strategies yielded the best detection performance (as shown in experiment ID 7).


\section{Conclusions}

In this work, we proposed a new method for face detection using deep learning techniques. Specifically, we extended the state-of-the-art Faster RCNN framework for generic object detection, and proposed several effective strategies for improving the Faster RCNN algorithm for resolving face detection tasks, including feature concatenation, multi-scale training, hard negative mining, and proper configuration of anchor sizes for RPN, etc.
We conducted an extensive set of experiments on the well-known FDDB testbed for face detection benchmark, and achieved the state-of-the-art results which ranked the best among all the published methods. Future work will further address the efficiency and scalability of the proposed method for real-time face detection.

\section*{Acknowledgments}

We would like to thank Hanfu Zhang for his suggestions and contributions to this project.



{
\bibliography{fddb_TR}
}

\end{document}